\documentclass[letterpaper]{article} 
\usepackage{aaai24}  
\usepackage{times}  
\usepackage{helvet}  
\usepackage{courier}  
\usepackage[hyphens]{url}  
\usepackage{graphicx} 
\usepackage{natbib}  
\usepackage{caption} 
\usepackage{algorithm}
\usepackage{algorithmic}

\usepackage{newfloat}
\usepackage{listings}

\usepackage{color}
\usepackage{amsfonts,amssymb,amsmath} 
\usepackage{xspace}
\usepackage{booktabs}
\newcommand{\themodel}{TGM-DLM\xspace}

\DeclareCaptionStyle{ruled}{labelfont=normalfont,labelsep=colon,strut=off} 
\floatstyle{ruled}
\newfloat{listing}{tb}{lst}{}
\floatname{listing}{Listing}
\title{Text-Guided Molecule Generation with Diffusion Language Model}
\author {
    Haisong Gong\textsuperscript{\rm 1,\rm 2},
    Qiang Liu\textsuperscript{\rm 1,\rm 2},
    Shu Wu\textsuperscript{\rm 1,\rm 2}\thanks{Corresponding Author},
    Liang Wang\textsuperscript{\rm 1,\rm 2}
}
\affiliations {
    \textsuperscript{\rm 1}Center for Research on Intelligent Perception and Computing\\
State Key Laboratory of Multimodal Artificial Intelligence Systems\\
Institute of Automation, Chinese Academy of Sciences\\
    \textsuperscript{\rm 2}School of Artificial Intelligence, University of Chinese Academy of Sciences\\
    gonghaisong2021@ia.ac.cn, \{qiang.liu, shu.wu, wangliang\}@nlpr.ia.ac.cn
}

\begin{document}

\maketitle

\begin{abstract}
Text-guided molecule generation is a task where molecules are generated to match specific textual descriptions. Recently, most existing SMILES-based molecule generation methods rely on an autoregressive architecture. In this work, we propose the Text-Guided Molecule Generation with Diffusion Language Model (\themodel), a novel approach that leverages diffusion models to address the limitations of autoregressive methods. \themodel updates token embeddings within the SMILES string collectively and iteratively, using a two-phase diffusion generation process. The first phase optimizes embeddings from random noise, guided by the text description, while the second phase corrects invalid SMILES strings to form valid molecular representations. We demonstrate that \themodel outperforms MolT5-Base, an autoregressive model, without the need for additional data resources. Our findings underscore the remarkable effectiveness of \themodel in generating coherent and precise molecules with specific properties, opening new avenues in drug discovery and related scientific domains. Code will be released at: https://github.com/Deno-V/tgm-dlm.
\end{abstract}

\section{Introduction} 

Molecules, the fundamental building blocks of matter, intricately shape the properties and functions of our world.  Novel molecules hold profound significance across scientific realms \cite{yao2016molecular,reiser2022graph,montoya2017high}, motivating research in fields like chemistry, materials science, and biology. Central to this pursuit is drug discovery, where identifying molecules with specific properties, particularly interactions with proteins or enzymes, is paramount \cite{ferreira2015molecular}.

Traditional drug discovery's resource-intensive nature is being transformed by artificial intelligence (AI) \cite{paul2021artificial}. Among AI's applications, generating drug-like molecules has drawn attention \cite{bagal2021molgpt,you2018graph,guan2023decompdiff}, along with bridging molecules and language \cite{zeng2022deep,liu2022multi}. In this paper, we focus on a novel task presented by \citet{edwards2022translation}: \emph{text-guided de novo molecule generation}. This innovative endeavor blends natural language with molecular structures, generating molecules based on textual descriptions. This task empowers more comprehensive control over molecule generation and transcends prior limitations in AI-assisted molecule design.

One widely adopted approach in the scientific community to represent a molecule is the simplified molecular-input line-entry system (SMILES) \cite{weininger1988smiles, weininger1989smiles}.  SMILES provides a compact and human-readable representation of chemical structures. As Figure \ref{figure:pre}(a) shows, atoms and bonds are represented by characters and symbols, enabling the encoding of complex molecular structures as strings. Recent research has extensively explored SMILES-based molecule generation, treating SMILES strings as a  form of language and applying natural language generation techniques to produce innovative molecules. 
Among these efforts, the majority of contemporary SMILES-based molecule generation methods rely on an autoregressive architecture, wherein models predict the next character based on previously generated ones \cite{bagal2021molgpt,frey2022neural,edwards2022translation,irwin2022chemformer}. Such methods often build upon autoregressive models like GPT \cite{floridi2020gpt}, T5 \cite{raffel2020exploring} and BART \cite{lewis2019bart}.

Despite their proven success, the autoregressive nature of existing methods brings forth inherent limitations, where the fixed generation order constrains the models' adaptability, particularly in settings that demand precise control over the generation process \cite{li2022diffusion}. \citet{bubeck2023sparks} demonstrates that autoregressive models, including the state-of-the-art GPT-4, encounter significant difficulties when facing content generation tasks under \textit{global constraints}. This arises because the autoregressive nature 
prevents the model from revising previously generated content, compelling it to focus on predicting content far ahead instead. Within the domain of SMILES-based molecule generation, the textual descriptions of molecules represent these crucial global constraints. These descriptions encapsulate vital molecular attributes, including scaffold structures, deeply embedded throughout the entire SMILES sequence. Consequently, the autoregressive architecture hinders accurate incorporation of crucial global constraints represented by textual description in SMILES-based molecule generation. This inherent limitation underscores the need for a novel generation paradigm that offers enhanced control.

To address the aforementioned nature of autoregressive nature, we explore the utilization of diffusion models in SMILES-based molecule generation. Unlike autoregressive models, diffusion models \cite{ho2020denoising} generate content iteratively and holistically. These models have exhibited remarkable aptitude in capturing complex data distributions and accommodating global constraints in the image generation field \cite{rombach2022high,yang2022diffusion}.  Inspired by these successes,  we present the Text-Guided Molecule Generation with Diffusion Language Model  (\textbf{\themodel}), a novel approach that leverages diffusion model to update token embeddings within the SMILES string collectively and iteratively.  In this process, \themodel executes a two-phase diffusion generation. In the first phase, \themodel iteratively optimizes the embedding features from random noise, guided by the text description. However, some molecules generated during this phase may suffer from invalid issues, with SMILES strings that cannot represent real molecules due to problems including unclosed rings, unmatched parentheses and valence errors. To address this, the second phase acts as a correction phase, where the model iteratively optimizes the invalid SMILES strings without text guidance, ultimately forming valid representations. To achieve these improvements, \themodel is trained with two objectives: denoising embeddings using the text description and recovering uncorrupted SMILES strings by deliberately feeding the model with corrupted invalid ones during the latter diffusion generation steps.

In a nutshell, our main contributions can be listed as follows,
\begin{itemize}
    \item We are the first to introduce diffusion language model to SMILES-based text-guided molecule generation, offering a powerful approach for coherent and precise molecule generation.
    \item We propose \themodel, a novel method with a two-phase diffusion generation process, enabling the generation of coherent molecules guided by text descriptions.
    \item \themodel showcases superior performance, notably surpassing MolT5-Base, an autoregressive generation model pretrained on the Colossal Clean Crawled Corpus (C4) \cite{raffel2020exploring} and ZINC-15 \cite{sterling2015zinc} dataset and fine-tuned on the ChEBI-20 dataset \cite{edwards2021text2mol}. Notably, these results are achieved without any additional data resources, highlighting the effectiveness of our model.
\end{itemize}

\section{Related Work}
\subsection{SMILES-based Molecule Generation}
Molecules, often analyzed computationally, can be represented in various formats, including SMILES, graphs, and 3D structures \cite{chen2023uncovering,zhu2022featurizations}. In the realm of SMILES-based molecule generation, early approaches employed RNN-based methods that transformed SMILES strings into one-hot vectors and were sampled step-by-step \cite{segler2018generating,grisoni2020bidirectional}. VAE-based methods, like ChemVAE \cite{gomez2018automatic} and SD-VAE \cite{dai2018syntax}, harnessed paired encoder-decoder architectures to generate SMILES strings and establish latent molecular spaces.

In recent years, the progress in natural language generation has sparked a surge in interest in autoregressive methods for molecule generation. Prominent models like GPT \cite{floridi2020gpt} have exhibited impressive capabilities. ChemGPT \cite{frey2022neural} and MolGPT \cite{bagal2021molgpt} leverage GPT-based architectures for molecule generation, with MolGPT focusing on generating molecules with specific attributes. Chemformer \cite{irwin2022chemformer}, on the other hand, employs BART \cite{lewis2019bart} as its foundational framework for molecule generation. MolT5 \cite{edwards2022translation}, closely aligned with our work, employs a pretrained T5 \cite{raffel2020exploring} to facilitate text-guided SMILES-based molecule generation. Distinctively, our approach stands out as the pioneer in utilizing the diffusion architecture as opposed to the autoregressive method.

\subsection{Diffusion Models for Language Generation}

Diffusion models have achieved notable accomplishments in generating content across continuous domains, spanning images \cite{ho2020denoising,rombach2022high} and audios \cite{kong2020diffwave}. Adapting diffusion models for discrete domains, such as language generation, has engendered an array of exploratory strategies. Some techniques incorporate discrete corruption processes, replacing the continuous counterpart, and leverage categorical transition kernels, uniform transition kernels, and absorbing kernels \cite{hoogeboom2021argmax,hoogeboom2021autoregressive,he2022diffusionbert}. In contrast, certain methodologies maintain diffusion steps within the continuous domain by transforming language tokens into word vectors, executing forward and reverse operations within these vectors \cite{li2022diffusion,gong2022diffuseq}. Our work adheres to the latter paradigm, operating on word vectors, and significantly broadening the capabilities of diffusion models within the realm of SMILES-based molecule generation. While the exploration of diffusion models for language generation remains relatively under-explored, our approach methodically investigates its potential within the SMILES-based molecule generation domain, effectively underscoring its effectiveness.

\section{Preliminary and Task Formulation}
\newcommand*{\emb}{\mathop{\mathrm{Emb}}}
\newcommand*{\attention}{\mathop{\mathrm{Attention}}}
\newcommand*{\softmax}{\mathop{\mathrm{softmax}}}
\newcommand*{\pe}{\mathop{\mathrm{PE}}}
\newcommand*{\de}{\mathop{\mathrm{DE}}}
\newcommand*{\mlp}{\mathop{\mathrm{MLP}}}
\newcommand*{\corrupt}{\mathop{\mathrm{Corrupt}}}
\subsection{Diffusion Framework for Language Generation}\label{section:pre}
\begin{figure}[tb]
\centering
\includegraphics[width=0.99\columnwidth]{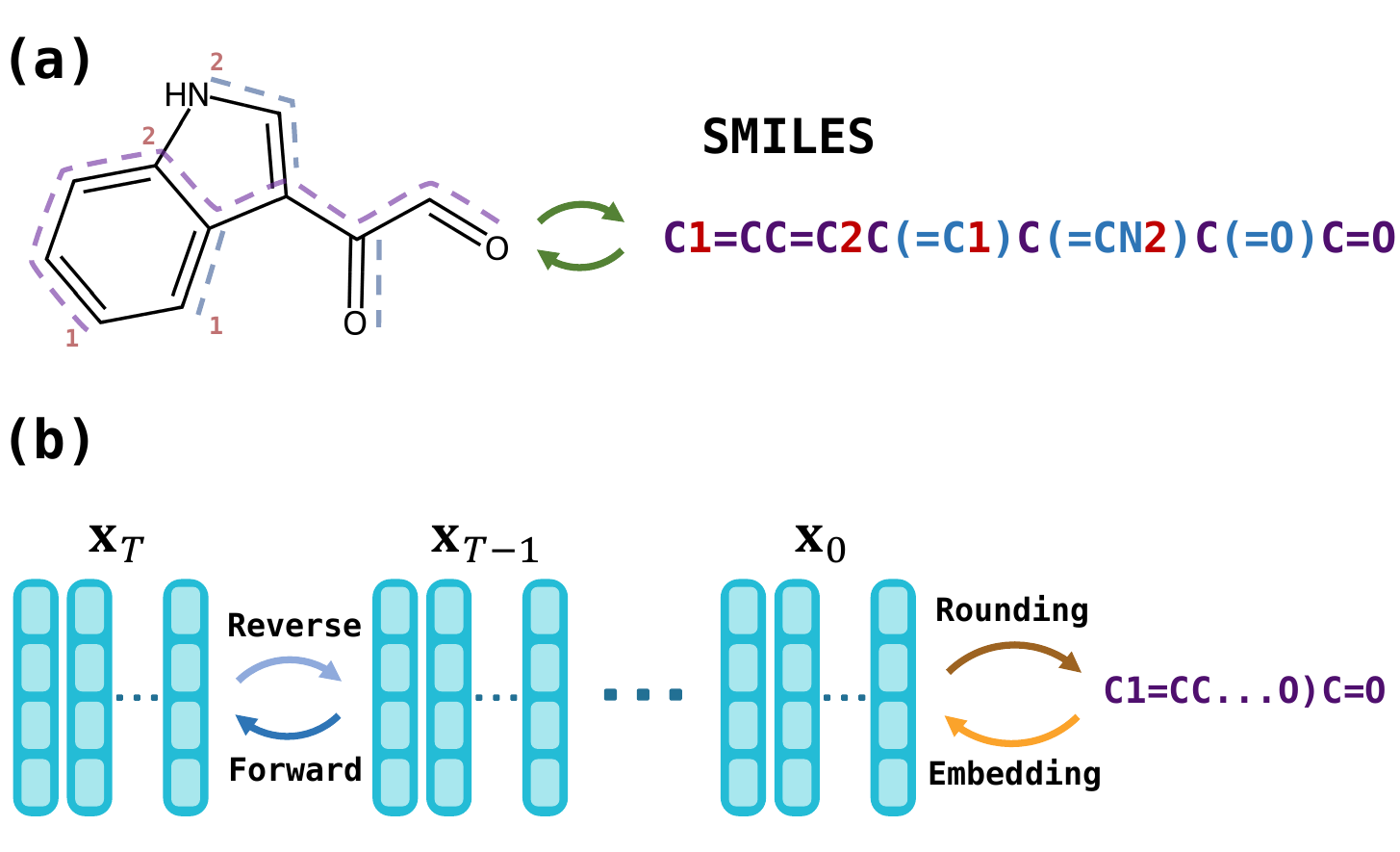} 
\caption{(a) Depiction of a molecule along with its corresponding SMILES representation.  The main chain and side chains are colored purple and blue, respectively, both in the molecule graph and SMILES string. Ring numbering is highlighted in red. (b) The fundamental framework of the diffusion model for language generation. SMILES is treated as a sequence of language tokens. Through embedding and forward processes, the sequence transforms into pure noise. The reverse and rounding processes reconstruct the SMILES string from pure noise.}
\label{figure:pre}
\end{figure}

In this section, we lay the foundation for our work by introducing a comprehensive framework for the application of diffusion models in continuous domains, modified from the work of \cite{li2022diffusion}.  The framework comprises four pivotal processes: embedding, forward, reverse, and rounding. Together, these processes enable the generation of coherent and meaningful language output, as depicted in Figure \ref{figure:pre}(b).

The embedding process is the initial step where a text sequence $W = [w_0, w_1, \ldots, w_n]$ is treated as a sequence of words. Each word $w_i$ undergoes an embedding transformation to yield $\emb(W) = [\emb(w_0), \emb(w_1), \ldots, \emb(w_n)] \in \mathbb{R}^{d \times n}$, with $n$ denoting sequence length and $d$ representing the embedding dimension. The starting matrix for the forward process $\mathbf{x}_0$ emerges by sampling from a Gaussian distribution centered at $\emb(W)$: $\mathbf{x}_0 \sim \mathcal{N}(\emb(W), \sigma_0 \mathbf{I})$.

The forward process gradually adds noise to $\mathbf{x}_0$, ultimately leading to the emergence of pure Gaussian noise $\mathbf{x}_T\sim\mathcal{N}(0,\mathbf{I})$. The transition from $\mathbf{x}_{t-1}$ to $\mathbf{x}_t$ is defined as follows:
\begin{equation}
 q(\mathbf{x}_t|\mathbf{x}_{t-1})=\mathcal{N}(\mathbf{x}_t;\sqrt{1-\beta_t}\mathbf{x}_{t-1},\beta_t\mathbf{I})   
\end{equation}
where $\beta_t\in[0,1]$ regulates the noise scale added during diffusion time step $t$, which is predefined constant.

The reversal process commences from $\mathbf{x}_T$ and progressively samples $\mathbf{x}_{t-1}$ based on $\mathbf{x}_{t}$, reconstructing the original content.  Traditionally, this is achieved through a neural network trained to predict $\mathbf{x}_{t-1}$ given $\mathbf{x}_t$. However, for enhanced precision in denoising $\mathbf{x}_t$ towards some specific word vectors, a neural network $f_\theta$ is trained to directly predict $\mathbf{x}_0$ from $\mathbf{x}_t$. In this way, the denoising transition used here from $\mathbf{x}_{t}$ to $\mathbf{x}_{t-1}$ can be formulated as:
\begin{equation}\label{formula:denoise}
\begin{split}
p_\theta(\mathbf{x}_{t-1}|\mathbf{x}_t)=\mathcal{N}(&\mathbf{x}_{t-1};\frac{\sqrt{\overline\alpha_{t-1}}\beta_t}{1-\overline\alpha_t}f_\theta(\mathbf{x}_t,t)\\
+&\frac{\sqrt{\alpha_t}(1-\overline\alpha_{t-1})}{1-\overline\alpha_t}\mathbf{x}_t,\frac{1-\overline\alpha_{t-1}}{1-\overline\alpha_t}\beta_t)
\end{split}
\end{equation}
where $\overline\alpha_t=\prod_{s=0}^t(1-\beta_s)$. Sequentially applying Equation \ref{formula:denoise} to a pure noise $\mathbf{x}_T$ enables us to iteratively sample $\mathbf{x}_{t-1}$, thereby yielding $\mathbf{x}_0$.

The rounding process completes the framework by reverting the embedding matrix to the original text sequence. Each column vector of $\mathbf{x}_0$ is mapped to the word with the closest L-2 distance in terms of word embeddings. Thus, through the reverse and rounding processes, any given noise can be effectively denoised to a coherent text output.

\subsection{Task Formulation}
The objective of text-guided molecule generation is to craft a molecule that aligns with a provided textual description. In a formal context, let $C = [w_0, w_1, \ldots, w_m]$ represent the given text description, where $w_i$ denotes the $i$-th word within the text, $m$ is the length of the text sequence. Our goal revolves around the construction of a model $\mathcal{F}$ that accepts the text as input and yields the intended molecule as output, mathematically expressed as $M = \mathcal{F}(C)$. 

\section{Method}
\subsection{Overview}

\begin{figure}[t]
\centering
\includegraphics[width=0.99\columnwidth]{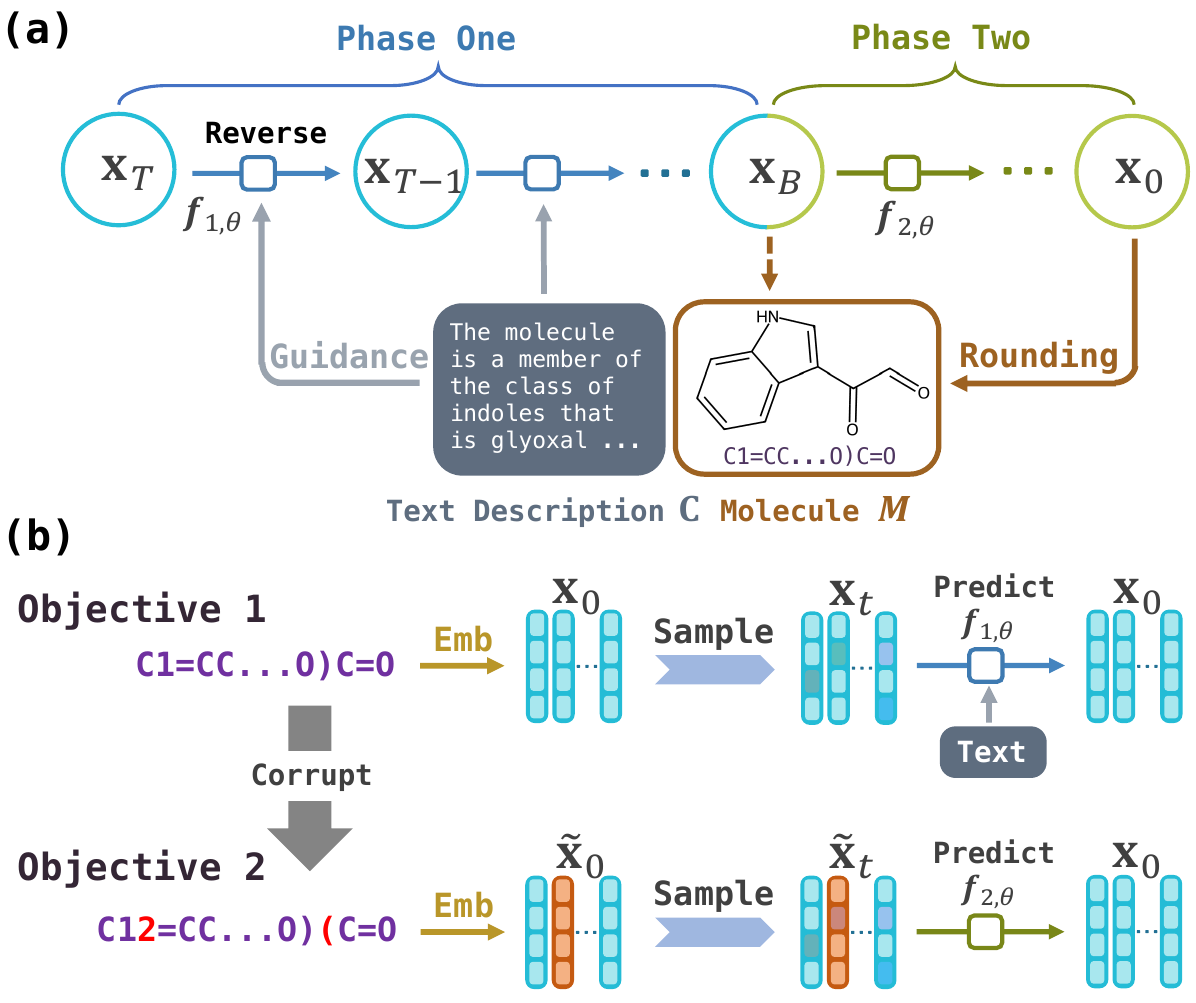} 
\caption{(a) Illustration of \themodel's two-phase diffusion process. Phase one starts from pure noise, denoising $\mathbf{x}_t$ to $\mathbf{x}_B$ under text guidance. Phase two, without guidance, corrects phase one's outputs that can't be rounded to valid SMILES strings. (b) Two training objectives designed for \themodel. The first objective entails denoising under text guidance, ensuring alignment with text descriptions. The second objective aims to enhance the model's ability to rectify invalid content, achieved by training it to recover embeddings from intentionally corrupted versions. }
\label{figure:main}
\end{figure}

Building upon the framework outlined in the preceding section, \themodel follows a similar structure, generating a molecule's SMILES string by iteratively denoising a pure Gaussian noise $\mathbf{x}_T$ through a two-phase reverse process, as depicted in Figure \ref{figure:main}(a). The first phase constitutes a reverse process encompassing $T-B$ denoising steps. It initiates with $\mathbf{x}_T$ and progressively refines the embedding matrix under the guidance of the text description $C$. Once the first phase concludes, an embedding matrix $\mathbf{x}_B$ is derived. This matrix is then scrutinized to verify if it corresponds to a valid molecule. The transformation of $\mathbf{x}_B$ to a SMILES string is facilitated by a rounding process, with the SMILES string's validity determined using the RDKit toolkit\footnote{https://rdkit.org}. Should $\mathbf{x}_B$ not correspond to any actual molecule, it proceeds to the second phase—known as the correction phase. In this phase, which also comprises $B$ denoising steps, $\mathbf{x}_B$ is employed as the starting point, ultimately culminating in the final embedding matrix $\mathbf{x}_0$. The $\mathbf{x}_0$ is then rounded to SMILES string, serving as the model's output.
In the upcoming sections, we delve into the specifics of each component of our design.

\subsection{SMILES Tokenizer}
Given our use of SMILES for molecular representation, each molecule $M$ is treated as a sequence. Though the simplest approach would be to tokenize the SMILES string by individual characters, this approach poses significant challenges. It disrupts the unity of multi-character units and introduces ambiguity. For instance, the atom ``scandium'' represented in SMILES as $[\text{Sc}]$ incorporates $\text{S}$ and $\text{c}$, representing sulfur and aromatic carbon, respectively. Moreover, numeric characters within SMILES can either denote ring numbers or exist within atom groups like $[\text{NH3+}]$, leading to ambiguity. Other methods, like those in autoregressive generation models, often rely on segmentation algorithms such as BPE \cite{gage1994new}. In this study, we advocate treating each atom and atom group as a unified token, including additional tokens for bonds, ring numbers, parentheses, and special cases, thereby forming our vocabulary.

As a result, for a given molecule $M$, it is represented as a sequence list $M=[a_0,a_1,\cdots,a_n]$, where $a_i$ represents the $i$-th token, and $n$ is the maximum sequence length. For instance, the SMILES $\text{C(C(=O)O)[NH3+]}$ is tokenized as $[\text{[SOS],C,(,C,(,=,O,),O,),[NH3+],[EOS],[PAD],...,[PAD]}]$. We pad every sequence to the maximum sequence length $n$.

After tokenization, the tokenized molecule $M$ is ready for the embedding process, resulting in $\mathbf{x}_0$ via the sampling of $\mathbf{x}_0\sim\mathcal{N}(\emb(M),\sigma_0\mathbf{I})$.

\subsection{Phase One: Text-Guided Generation}
The primary stage of diffusion generation encompasses a complete process involving $T-B$ diffusion steps. In contrast to generating molecules directly from unadulterated noise, the first phase of \themodel incorporates text guidance to shape its generation process.

Several strategies exist to infuse the reverse process with the textual context. For instance, \citet{li2022diffusion} employs an auxiliary classifier to steer the reverse process. Additionally, \citet{rombach2022high} enhanced diffusion models to serve as more versatile image generators, capable of being directed by semantic maps, textual descriptions, and images. This augmentation involves integrating the underlying model with a cross-attention mechanism. Drawing inspiration from this advancement, we incorporate cross-attention mechanism within \themodel.

 To introduce the text description, we utilize a pre-trained language model to map the text sequence $C$ to its latent embeddings $\mathbf{C}\in \mathbb{R}^{d_1\times m}$, where $d_1$ denotes the output embedding dimension of the language model. Our methods employs a Transformer model as the backbone, powering the neural network $f_{\theta}$ in Equation \ref{formula:denoise}.  We denote the function as $f_{1,\theta}$ to emphasize that it is used for the reverse process of phase one. Given the current diffusion state $\mathbf{x}_t$, diffusion step $t$, and text embeddings $\mathbf C$, $f_{1,\theta}$ predicts $\mathbf{x}_0$ as $\hat{\mathbf{x}}_0=f_{1,\theta}(\mathbf{x}_t,t,\mathbf C)$. We encode diffusion step $t$ into the input of the first Transformer layer $\mathbf{z}_{t}^{(0)}$ using a technique akin to positional embedding \cite{vaswani2017attention}:
 \begin{equation}
\mathbf{z}_{t}^{(0)} = \mathbf{W}_{in}\mathbf{x}_t  + \mathrm{PosEmb} + \de(t)
 \end{equation}
 where $\de$ transforms $t$ a vector, $\mathbf{W}_{in}\in\mathbb{R}^{d_2\times d}$, $\mathbf{z}_{t}^{(0)}\in\mathbb{R}^{d_2\times n}$, $d_2$ is the dimension for Transformer. $\mathrm{PosEmb}$ stands for the positional embedding.

The Transformer comprises $L$ layers, each layer contains a cross-attention block, which introduces the text description into hidden states through the following mechanism:
\begin{equation}
\begin{split}
\attention(\mathbf{Q},\mathbf{K},\mathbf{V})&=\mathbf{V}\softmax(\frac{\mathbf{K}^T\mathbf{Q}}{\sqrt{d_2}})
\\
\mathbf{Q}&=\mathbf{W}_Q^{(i)}\mathbf{z}_t^{(i)}
\\
\mathbf{K}&=\mathbf{W}_K^{(i)}\mlp(\mathbf{C})
\\ 
\mathbf{V}&=\mathbf{W}_V^{(i)}\mlp(\mathbf{C})
\end{split}
\end{equation}
where, $\mathbf{W}_*^{(i)}\in\mathbb{R}^{d_2\times d_2}$ represents the learnable parameters of the cross-attention block within the $i$-th layer, $\mlp$ is a multilayer perceptron.

After the process of phase one, we get the matrix $\mathbf{x}_B$ from pure noise $\mathbf{x}_T$ under the guidance of text description $C$. $\mathbf{x}_B$ is ready for the rounding process and being converted to SMILES string.

\subsection{Phase Two: Correction}
Following the diffusion process of phase one, the resulting SMILES strings obtained from $\mathbf{x}_B$ through the rounding process might occasionally fail to constitute a valid string. This underscores the need for a secondary phase—phase two—dedicated to rectifying such instances. As depicted in Figure \ref{figure:pre}(a), SMILES strings adhere to specific grammatical rules involving parentheses and numbers. Paired parentheses indicate branching on the main chain, while matching numbers represent bonds between atoms. To be valid, SMILES strings must satisfy these criteria, including meeting valence requirements for each atom.

By our observation, about three-quarters of the invalid SMILES strings generated by phase one fail to have paired parentheses and numbers. These erroneous SMILES strings have absorbed ample information from the text description, yet they contain minor inaccuracies in terms of rings and parentheses. Such issues can typically be rectified through minor adjustments involving the addition or removal of numbers and parentheses.

Unlike the condition-driven diffusion process of phase one, phase two's architecture closely mirrors its predecessor, employing $B$ diffusion steps without text guidance, as adequate text information has already been acquired.  Operating on the Transformer framework, phase two employs the denoising network $f_{2,\theta}$ as a post-processing module to refine $\mathbf{x}_B$ into a valid $\mathbf{x}_0$. It's important to note that this correction process introduces some distortion, affecting original information. Balancing effective correction with controlled distortion is achieved by tuning the steps taken during phase two, a topic explored in the upcoming experimental section.

\subsection{Training}
To endow the denoising network with the capability to execute both phase one and phase two, we equip \themodel with two training objectives, as depicted in Figure \ref{figure:main}(b).

\subsubsection{Objective One: Denoising Training} For the training of phase one, our approach centers around maximizing the variational lower bound of the marginal likelihood. Simplified and adapted from the works of \citet{ho2020denoising} and \citet{li2022diffusion}, the objective function compels $f_{1,\theta}$ to recover $\mathbf{x}_0$ at each step of the diffusion process:
\begin{equation}
\begin{aligned}
\mathcal{L}_1(M,C)=\mathop{\mathbb{E}}_{q(\mathbf{x}_{0:T}|M)}
\Big[
\sum_{t=1}^{T}\|f_{1,\theta}&(\mathbf{x}_t,t,\mathbf{C})-\mathbf{x}_0\|^2
\\
&-\log p_\theta(M|\mathbf{x}_0)
\Big]
\end{aligned}
\end{equation}
where $p_\theta(M|\mathbf{x}_0)$ signifies the rounding process, characterized by a product of softmax distribution $p_\theta(M|\mathbf{x}_0)=\prod_{i=0}^n p(a_i|\mathbf{x}_{0[:,i]})$.

\subsubsection{Objective Two: Corrective Training} For the training of phase two, our approach deviates from predicting $\mathbf{x}_0$ from $\mathbf{x}_t$ and $\mathbf{C}$; instead, we compel the network to forecast $\mathbf{x}_0$ from a corrupted variant, denoted as $\Tilde{\mathbf{x}}_t$, stemming from $\mathbf{x}_t$. As the primary objective of phase two revolves around rectifying unmatched rings and parentheses issues, we deliberately introduce such irregularities into $\mathbf{x}_t$.

In a conventional diffusion model training scenario, the acquisition of $\mathbf{x}_t$ involves the determination of $\mathbf{x}_0$ and the diffusion step $t$, followed by sampling via the equation:
\begin{equation}
\mathbf{x}_t = \sqrt{\overline\alpha_t}\mathbf{x}_0+\sqrt{1-\overline\alpha_t}\epsilon    
\end{equation}
where $\epsilon\sim\mathcal{N}(0,\mathbf{I})$. 

To sample the corrupted $\Tilde{\mathbf{x}}_t$, we initially infuse unmatched ring and parentheses complexities into the original molecule sequence $M$. This is achieved through a function $\corrupt$:
\begin{equation}\label{equation:corrupt}
\Tilde{M} = \corrupt(M)    
\end{equation}
resulting in the altered molecule $\Tilde{M}$.  The function $\corrupt$ incorporates a probability $p$ of randomly adding or removing varying numbers of parentheses and ring numbers to disrupt the paired instances. Then, we transform $\Tilde{M}$ to $\Tilde{\mathbf{x}}_0$ through the embedding process, and $\Tilde{\mathbf{x}}_t$ is sampled by:
\begin{equation}\label{equation:cor_sample}
\Tilde{\mathbf{x}}_t = \sqrt{\overline\alpha_t}\Tilde{\mathbf{x}}_0+\sqrt{1-\overline\alpha_t}\epsilon    
\end{equation}
The training loss for phase two becomes:
\begin{equation}
\begin{aligned}
\mathcal{L}_2(&M,\mathbf{C})=\mathop{\mathbb{E}}_{q(\mathbf{x}_{0:T}|M)}
\Big[
\sum_{t=1}^{\tau}\|f_{2,\theta}(\Tilde{\mathbf{x}}_t,t)-\mathbf{x}_0\|^2
\\
&
+\sum_{t=\tau}^{T}\|f_{2,\theta}(\mathbf{x}_t,t,\mathbf{C})-\mathbf{x}_0\|^2
-\log p_\theta(M|\mathbf{x}_0)
\Big]
\end{aligned}
\end{equation}

Note that when training $f_{2,\theta}$, we maintain the first $T-\tau$ reverse process steps in alignment with phase one training. The introduction of corruption is confined only to the final $\tau$ steps. This strategic placement accounts for the fact that in the initial stages of the reverse process, $\mathbf{x}_t$ retains limited information, resembling noise, and any corruption introduced at this point could impede training and hinder convergence. See Algorithm \ref{alg} for the complete training procedure.

\begin{algorithm}[tb]
\caption{Training algorithm}
\label{alg}
\begin{algorithmic}[1] 
\REPEAT
\STATE Sample $M$, $\mathbf{C}$ and $t$. 
\STATE Obtain $\mathbf{x}_0$ by embedding process
\IF {$t=0$}
\STATE $g=\nabla_\theta \left(-\log p_\theta(M|\mathbf{x}_0)\right)$
\ELSIF {phase two \textbf{and} $t<\tau$}
\STATE Obtain $\Tilde{x}_t$ by Equation \ref{equation:corrupt} and \ref{equation:cor_sample}
\STATE $g=\nabla_\theta\|f_{2,\theta}(\Tilde{\mathbf{x}}_t,t)-\mathbf{x}_0\|^2$
\ELSE
\STATE $g=\nabla_\theta\|f_{*,\theta}(\mathbf{x}_t,t,\mathbf{C})-\mathbf{x}_0\|^2$
\ENDIF
\STATE Take one step of optimization through gradient $g$
\UNTIL{converged}
\end{algorithmic}
\end{algorithm}

\begin{table*}[t]
  \centering
  \tabcolsep=4.2pt
  \resizebox{1\textwidth}{!}{
    \begin{tabular}{c|c|c|c|c|c|c|c|c|c}
    \toprule[2pt]
    Model & BLEU$\uparrow$  & Exact$\uparrow$ & Levenshtein$\downarrow$ & MACCS FTS$\uparrow$ & RDK FTS$\uparrow$ & Morgan FTS$\uparrow$ & FCD$\downarrow$ & Text2Mol$\uparrow$ & Validity$\uparrow$ \\
    \midrule
    Ground Truth & 1.000 & 1.000 & 0.000 & 1.000 & 1.000 & 1.000 & 0.00 & 0.609 & 1.000 \\
    Transformer  & 0.499 & 0.000 & 57.660 & 0.480 & 0.320 & 0.217 & 11.32 & 0.277 & \textbf{0.906} \\
    T5-Base& 0.762 & 0.069 & 24.950 & 0.731 & 0.605 & 0.545 & 2.48 & 0.499 & 0.660 \\
    MolT5-Base & 0.769 & \underline{0.081} & 24.458 & 0.721 & 0.588 & 0.529 & 2.18  & 0.496 & 0.772 \\
    $\text{\themodel}_{\text{w/o\ corr}}$& \textbf{0.828} & \textbf{0.242} & \textbf{16.897} & \textbf{0.874} & \textbf{0.771} & \textbf{0.722} & \underline{0.89} & \textbf{0.589} & 0.789 \\
    \themodel & \underline{0.826} & \textbf{0.242} & \underline{17.003} & \underline{0.854} & \underline{0.739} & \underline{0.688} & \textbf{0.77} & \underline{0.581} & \underline{0.871} \\
    \bottomrule[1.2pt]
    \end{tabular}%
}
\caption{Text-guided molecule generation results on ChEBI-20 test split. The results for Transformer, T5-Base and MolT5-Base are retrieved from \cite{edwards2022translation}. We bold the best scores and underline the second-best scores. $\text{\themodel}_{\text{w/o\ corr}}$ is the results generated by phase one, while \themodel combines two phases, with the second phase increasing Validity by 8.2\%.} 
  \label{tab:main}%
\end{table*}%

\section{Experiment}
In this section, we undertake experiments to assess the performance of our proposed \themodel in text-guided molecule generation. Additionally, we delve into the impact of the second phase within our two-phase generation framework, a different training approach, and the outcome of integrating text guidance during the correction phase.
\subsection{Experimental Setups}
\subsubsection{Dataset}
Given the nascent nature of our research focus, our evaluation centers on the ChEBI-20 dataset \cite{edwards2022translation}, which is currently the sole publicly available dataset. This dataset encompasses a collection of 33,010 molecule-description pairs, which are separated into 80/10/10\% train/validation/test splits. We adhere the data split setting in this paper.
\subsubsection{Metrics}
Following previous work \cite{edwards2022translation}, we employ nine metrics to evaluate the models.
\begin{itemize}
    \item \textbf{SMILES BLEU} score and \textbf{Levenshtein} distance.\ 
    Measure the similarity and distance of generated SMILES string to the ground truth molecule SMILES string. 
    \item \textbf{MACCS FTS}, \textbf{RDK FTS} and \textbf{Morgan FTS}.\ 
    Evaluate average Tanimoto similarity between generated and ground truth fingerprints.
    \item \textbf{Exact} score and \textbf{Validity}.\ 
    The percentage of generated molecules that exactly match the ground truth and the percentage of generated strings that are valid.
    \item \textbf{FCD} and \textbf{Text2Mol} score.\ 
    Measure latent information agreement using Fréchet ChemNet Distance (FCD) and assess relevance between text description and generated molecule using Text2Mol \cite{edwards2021text2mol}.
\end{itemize}

\subsubsection{Baselines}
Three baseline models are selected for comparison. All three models are autoregressive generation models.
\begin{itemize}
    \item \textbf{Transformer} \cite{vaswani2017attention}. A vanilla Transformer model with six encoder and decoder layers directly trained on the ChEBI-20 dataset. 
    \item \textbf{T5-Base} \cite{raffel2020exploring}. A sequence-to-sequence generation model pre-trained on the C4 dataset, fine-tuned on ChEBI-20 dataset.
    \item \textbf{MolT5-Base} \cite{edwards2022translation}. This model is initialized from a pre-trained T5 model, and further pre-trained on a combined dataset of C4 and ZINC-15 to gain domain knowledge in both molecules and natural languages. It is then fine-tuned on the ChEBI-20 dataset.
\end{itemize}
It's worth noting that the options for baseline models in the realm of text-based molecule generation task are limited. We choose the base versions of the T5 model and MolT5 model for a fair comparison, as language model's performance is positively correlated with the size of its parameters. The base version of T5/MolT5 comprises approximately 220M parameters, which is around 22\% larger than \themodel with about 180M parameters.

\subsection{Implementation Details}

We set the maximum sequence length for tokenized SMILES strings to $n=256$. Molecule-description pairs with SMILES string lengths exceeding 256 were filtered out (approximately 1\% of the entire dataset).  SMILES vocabulary contained 257 tokens, with trainable token embeddings set at $d=32$.We employed SciBERT \cite{beltagy2019scibert}
as our frozen encoder for text descriptions, with an embedding dimension of $d_1=768$. The Transformer network for $f_{*,\theta}$ comprises $L=12$ layers, and the hidden dimension is configured as $d_2=1024$. \themodel is composed of approximately 180M trainable parameters.

During molecule generation, we adopt a uniform skipping strategy for reverse steps to enhance sampling efficiency. As a result, the practical number of sample steps is 200 for phase one and 20 for phase two. Note that the steps used for phase two are between 0 to $\tau$.  In this way, it only takes about 1.2 seconds on average to generate one molecule from its description on our hardware (AMD EPYC 7742 (256) @ 2.250GHz CPU and one NVIDIA A100 GPU). 

During training, we set the total diffusion steps to $T=2,000$ for both phase one and phase two. For phase two, $\tau$ is set to 400, and the corruption probability $p$ is set to 0.4. We used Adam \cite{kingma2015adam} as the optimizer, employing linear warm-up and a learning rate of 1e-4. We train separate denoising models for each phase using objective one and objective two, respectively. It is noteworthy that the two denoising models share the same architecture. We also experimented with training a single model using the second objective for both phases, and we present the results of this comparison in the subsequent section.

\begin{figure*}[t]
\centering
\includegraphics[width=0.97\textwidth]{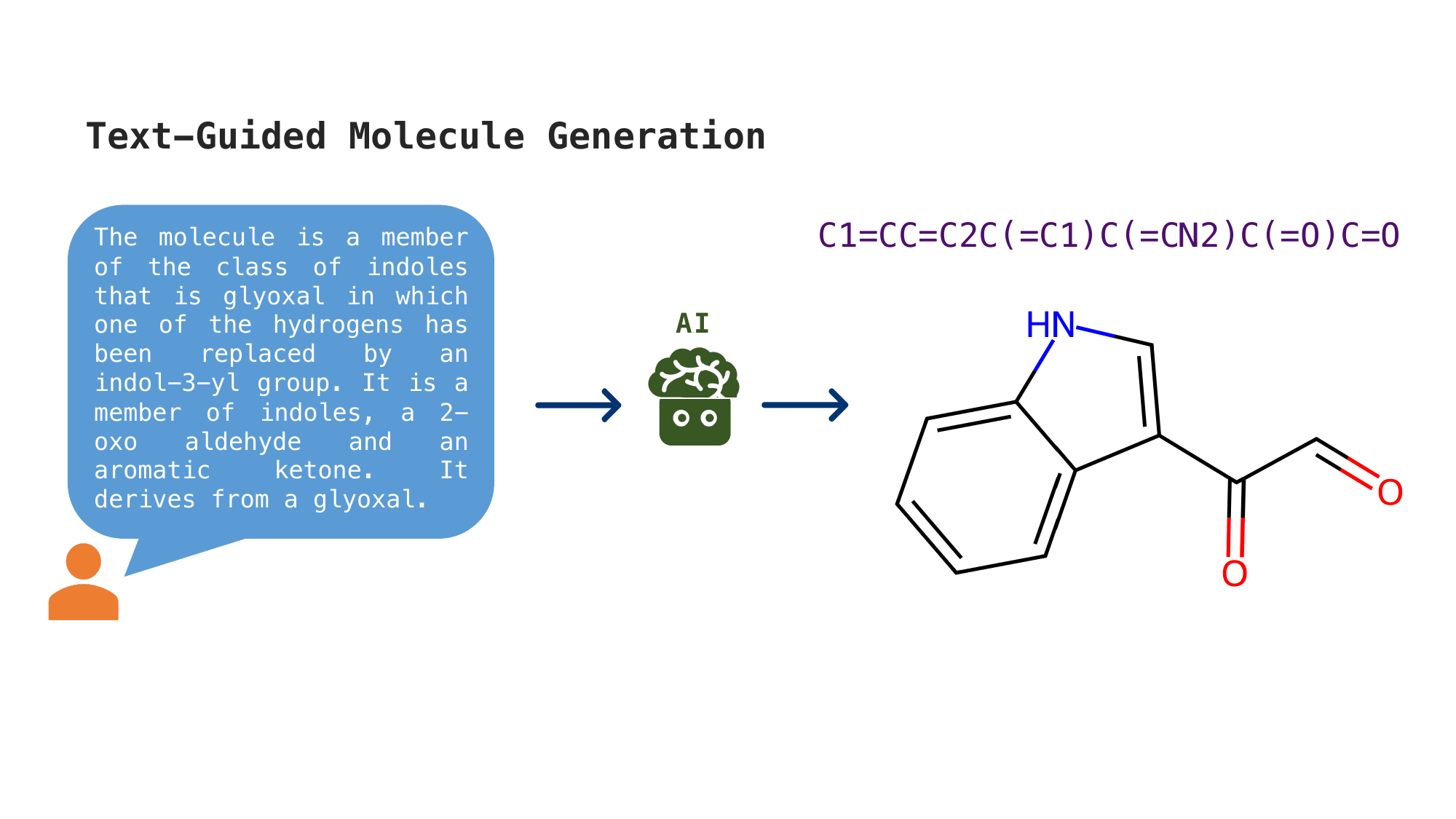} 
\caption{Example of molecules generated by different models with the same input descriptions. Generated SMILES strings are converted to molecule graphs for better visualization. }
\label{fig:examplemolecule}
\end{figure*}

\subsection{Overall Performance}
We compare \themodel with baseline models in Table \ref{tab:main}. In general, our model outperforms all baseline models. While Transformer excels in Validity, its performance is poor in other metrics, highlighting limited efficacy. In contrast, our model excels in all other metrics.  When compared to MolT5-base, our model exhibits a remarkable tripling of the exact match score, 18\% to 36\% improvement in FTS metrics. \themodel generates molecules that are most similar to the ground truth molecules and exhibit the closest alignment to the text descriptions, as evidenced by the highest Text2Mol score.  Notably, our model achieves these results without additional data or pre-training, distinguishing it from MolT5-Base. For more intuitive comparison, we showcase examples in Figure \ref{fig:examplemolecule}.

When comparing \themodel with $\text{\themodel}_{\text{w/o\ corr}}$, the latter represents results generated solely by phase one. We observe the successful rectification of invalid SMILES strings by phase two, boosting the Validity metric from 78.9\% to 87.1\%. However, this enhancement is accompanied by a slight reduction in other metrics. This can be attributed to two factors: firstly, the computation of metrics excluding BLEU, Exact, Levenshtein, and Validity considers only valid SMILES, thus the correction phase expands the metric scope. Secondly, the corrective phase inherently introduces subtle distortions in original molecular information, leading to a slight metric reduction.


\begin{table}[tbp]
  \centering
  \tabcolsep=3.3pt
  \resizebox{1\columnwidth}{!}{
    \begin{tabular}{c|c|c|c|c}
    \toprule[2pt]
    Model &BLEU&MACCS FTS&Text2Mol&Validity\\
    \midrule[1.07pt]
    $\text{\themodel}_{\text{w/o\ corr}}$& \textbf{0.828} & 
    \textbf{0.874} & \textbf{0.589} & 0.789\\
    \midrule
    $\text{\themodel}_{0.5\times}$ & \underline{0.827} & 0.858 & 0.584 & 0.855 \\
   $\text{\themodel}_{1\times}$ & 0.826 & 0.854 & 0.581 & 0.871 \\
    $\text{\themodel}_{2\times}$ & 0.825 & 0.851 & 0.580 & \underline{0.875} \\
    $\text{\themodel}_{3\times}$ & 0.825 & 0.849 & 0.579 & \textbf{0.883} \\
    $\text{\themodel}_{joint}$ & 0.819 & 0.846 & 0.576 & 0.855  \\
    $\text{\themodel}_{text}$ & \underline{0.827} & \underline{0.869} & \underline{0.588} & 0.824  \\
    \bottomrule[1.2pt]
    \end{tabular}%
}
\caption{Comparison of \themodel and its variants. Representative metrics are selected to display.  $\text{\themodel}_{2\times}$ is a variant with twice the number of phase two steps. $\text{\themodel}_{joint}$ denotes the variant using a singular network trained for both phases. $\text{\themodel}_{text}$ is a variant that incorporates text input in the correction phase. }
  \label{tab:sidetable}%
\end{table}%

\subsection{Influence of Correction Phase}
In this section, we investigate the impact of the two-phase design. A key factor is the number of steps employed during molecule sampling in phase two. As previously elucidated in the implementation section, we adopt a relatively short diffusion process for phase two. By varying the number of steps in phase two, we generate diverse variants of \themodel. For instance, $\text{\themodel}_{2\times}$ employs twice the number of phase two steps. The findings are presented in Table \ref{tab:sidetable}.
In general, validity increases as the number of phase two steps expands, albeit at the expense of other metrics. Given that only invalid SMILES strings from phase one are affected, adjusting the phase two step count allows us to flexibly strike a balance between Validity and the preservation of original information in erroneous SMILES.

\subsection{Joint Training of Two Phases}
As described in the implementation section, we separately train $f_{1,\theta}$ and $f_{2,\theta}$ with corresponding training objectives. Nevertheless, owing to the identical architecture of the two networks, we explore training a singular network for both phase one and phase two. This is achieved by employing Algorithm \ref{alg} while disregarding phase two condition at line 6. The results in Table \ref{tab:sidetable} indicate that $\text{\themodel}_{joint}$ performs less effectively than \themodel. A potential explanation is that the correction training objective could influence the regular training objective. 

\subsection{Correction with Text Input}
In \themodel, the correction phase operates independently of text input. However, we explore an alternative approach where correction is performed with the utilization of text descriptions as input. The outcomes of this approach are presented in Table \ref{tab:sidetable}. From the table we can see that employing text input for correction does not yield the expected significant improvement in validity. In fact, our experiments reveal that augmenting the number of diffusion steps in phase two for $\text{\themodel}_{text}$ fails to increase the validity beyond 82.4\%. These findings indicate that $\text{\themodel}_{text}$'s design is not well-suited for the correction phase, as the presence of text input may potentially impede the model's corrective ability by favoring adherence to textual guidance.

\section{Conclusion}
In this work, we present \themodel, a novel diffusion model for text-guided SMILES-based molecule generation. This approach embeds SMILES sequences and employs a two-phase generation process. The first phase optimizes embeddings based on text descriptions, followed by a corrective second phase to address invalid SMILES generated by the first phase.  Extensive experiments on ChEBI-20 dataset show \themodel's consistent outperformance of autoregressive baselines, demonstrating better overall molecular attributes understanding, notably tripling the exact match score and 18\% to 36\% improvement in fingerprinting metrics over MolT5-Base. Remarkably, these advancements are attained without additional data sources or pre-training, highlighting \themodel's effectiveness in text-guided molecule generation.

\section{Acknowledgements}
This work is jointly sponsored by National Natural Science Foundation of China (62206291, 62236010, 62141608).

\bibliography{aaai24}

\end{document}